\documentclass[runningheads]{llncs}
\usepackage[T1]{fontenc}
\usepackage{graphicx} 
\usepackage{multirow}
\usepackage{tabulary}
\usepackage{array}
\usepackage{soul}
\usepackage{xcolor}
\usepackage{footmisc}
\usepackage{bbding}

\newcolumntype{K}[1]{>{\centering\arraybackslash}p{#1}}
\newcolumntype{N}{>{\mbox}c}

\begin{document}

\title{Visionerves: Automatic and Reproducible  
\\Hybrid AI for
Peripheral Nervous System Recognition 
Applied to Endometriosis Cases}

\author{Giammarco La Barbera\inst{1,2(}\Envelope\inst{)}\and Enzo Bonnot\inst{3,2,1}\and Thomas Isla\inst{1}\and \\Juan Pablo de la Plata\inst{1,2}\and Joy-Rose Dunoyer de Segonzac\inst{1}\and Jennifer Attali\inst{4}\and Cécile Lozach\inst{4}\and Alexandre Bellucci\inst{5}\and  Louis Marcellin\inst{6,7} \and Laure Fournier\inst{5}\and Sabine Sarnacki\inst{8,1,2}\and Pietro Gori\inst{3,2} \and Isabelle Bloch\inst{9,3,1,2}}


\institute{IMAG2, Institut Imagine, Université Paris Cité, France \and Replico SAS, Paris, France \and LTCI, Télécom Paris, Institut Polytechnique de Paris, France\and  Université Paris Cité, Department of Pediatric Imaging, Hôpital Necker Enfants-Malades, Assistance Publique-Hôpitaux de Paris (AP-HP), France\and Université Paris Cité, Department of Radiology, Hôpital Européen Georges Pompidou, AP-HP, France\and Université Paris Cité, Department of Gynecological Surgery and Oncology (Professor Chapron), Hôpital Cochin, AP-HP, France\and 
Department of Development, Reproduction and Cancer (Professor Batteux), Institut Cochin, Paris, France\and
Université Paris Cité, Department of Pediatric Surgery, Hôpital Necker Enfants-Malades, AP-HP, France\and
Sorbonne Université, CNRS, LIP6, Paris\\
\email{giammarco.labarbera@replico.tech}
}

\authorrunning{G. La Barbera et al.}
\titlerunning{Visionerves}
\date{May 2025}

\maketitle

\begin{abstract}
Endometriosis often leads to chronic pelvic pain and possible nerve involvement, yet imaging the peripheral nerves remains a challenge. We introduce Visionerves, a novel hybrid AI framework for peripheral nervous system recognition from multi-gradient DWI and morphological MRI data. Unlike conventional tractography, Visionerves encodes anatomical knowledge through fuzzy spatial relationships, removing the need for selection of manual ROIs. The pipeline comprises two phases: (A) automatic segmentation of anatomical structures using a deep learning model, and (B) tractography and nerve recognition by symbolic spatial reasoning. Applied to the lumbosacral plexus in 10 women with (confirmed or suspected) endometriosis, Visionerves demonstrated substantial improvements over standard tractography, with Dice score improvements of up to 25\% and spatial errors reduced to less than 5 mm. This automatic and reproducible approach enables detailed nerve analysis and paves the way for non-invasive diagnosis of endometriosis-related neuropathy, as well as other conditions with nerve involvement.
\end{abstract}


\section{Introduction}
\label{sec:intro}
Endometriosis is a prevalent gynecological disease characterized by endometrial tissue outside the uterus, leading to chronic inflammation, immune dysfunction, and potential neurological involvement in the pelvic region. These neurological factors are difficult to diagnose with conventional imaging. Moreover, a better understanding of how pelvic nerve fibers are affected in endometriosis could help explaining the role they play in chronic pelvic pain~\cite{Chapron2019,Fauconnier2005}. Advanced techniques such as Diffusion Weighted MRI (DWI) and tractography~\cite{Tournier} enable visualization of nerve fibers, but their application has mainly been limited to the central nervous system (CNS)~\cite{Garyfallidis,Smith2012,Wassermann}, with limited studies in the peripheral nervous system (PNS)~\cite{Cage2015,Muller} and thus on endometriosis~\cite{Manganaro2014,Zijta2012}. This is due to: (i) the challenge of accurately and reproducibly reconstructing PNS via tractography algorithms, primarily due to reliance on manual placement of regions of interest (ROIs) and significant inter-subject variability; (ii) the complexity of recognizing each individual nerve bundle, due to the abundance of streamlines and spurious fibers (such as muscle, tissue and noise)~\cite{Haakma,Lemos}.

In this paper, we introduce ``Visionerves", an original hybrid AI method that leverages anatomical knowledge for automatic nerve identification, eliminating the need for manual ROIs. By describing nerve trajectories relatively to other anatomical structures (segmented via deep learning from a standard MRI image) and modeling anatomical imprecision with fuzzy logic~\cite{BlochRalescu,Hudelot}, our framework allows 
guiding the tractography algorithm, making the results more reproducible, and filtering out spurious (i.e. outliers) fibers by recognizing only nerve ones, via symbolic spatial reasoning. 
Furthermore, using a software such as 3DSlicer~\cite{Fedorov2012Slicer}, Visionerves results (anatomical segmentation and nerve fibers) can be merged and rendered into 3D models, offering valuable insights into the relationships between lesions, surrounding organs, and nerve structures~\cite{Vinit2024}. This approach could help explain the neuropathic component of chronic pain and paves the way for significant improvements in surgical and clinical planning. The presented method, applied in the pelvic region (from L5 to S3 nerve fibers) to 10 female adult subjects affected by endometriosis, yielded promising results, showing a good correlation with nerve reference reconstructions. 


\section{Related work}
\label{sec:sota}
Traditional methods employ the virtual dissection technique which requires manual placement of ROIs to select or exclude fibers~\cite{Haakma,Lemos,Manganaro2014}, but this process is labor-intensive, time-consuming, and difficult to reproduce for complex tracts. Atlas-based approaches, commonly used in the brain, transfer labels via non-linear deformations and mappings~\cite{Tournier2011}, but the accuracy of the results depends on alignment and clustering quality, which is problematic in pathological cases~\cite{Garyfallidis,Gori2014}.
Machine learning and deep learning methods, explored for the CNS~\cite{Knoedler2015,Wasserthal2018}, require large annotated datasets, are often hard to interpret, and can be affected by data biases such as site-specific protocol or scanner differences. Moreover, they do not address the inherent vagueness of PNS tract definitions, where boundaries are inherently ambiguous and difficult to delineate.

Taking a different approach, WMQL~\cite{Wassermann} introduced a query language for defining brain white matter tracts using mathematical models of spatial relationships and logical operations. However, the method is limited to binary relations, rough representation of structures, and does not account for pathological deviations. A fuzzy set theory extension~\cite{Delmonte} addresses the inherent vagueness of anatomical definitions, but remains tailored to CNS fibers, not accounting for the more complex spatial relationships seen in the PNS. For these reasons, these methods offer valuable insights, but are insufficient to reconstruct and recognize convoluted and smaller nerves of the PNS, such as the nerves of the pelvic lumbosacral plexus, which are clinically significant in conditions like endometriosis.

\begin{figure}[h!]
    \centering
    \includegraphics[width=\textwidth]{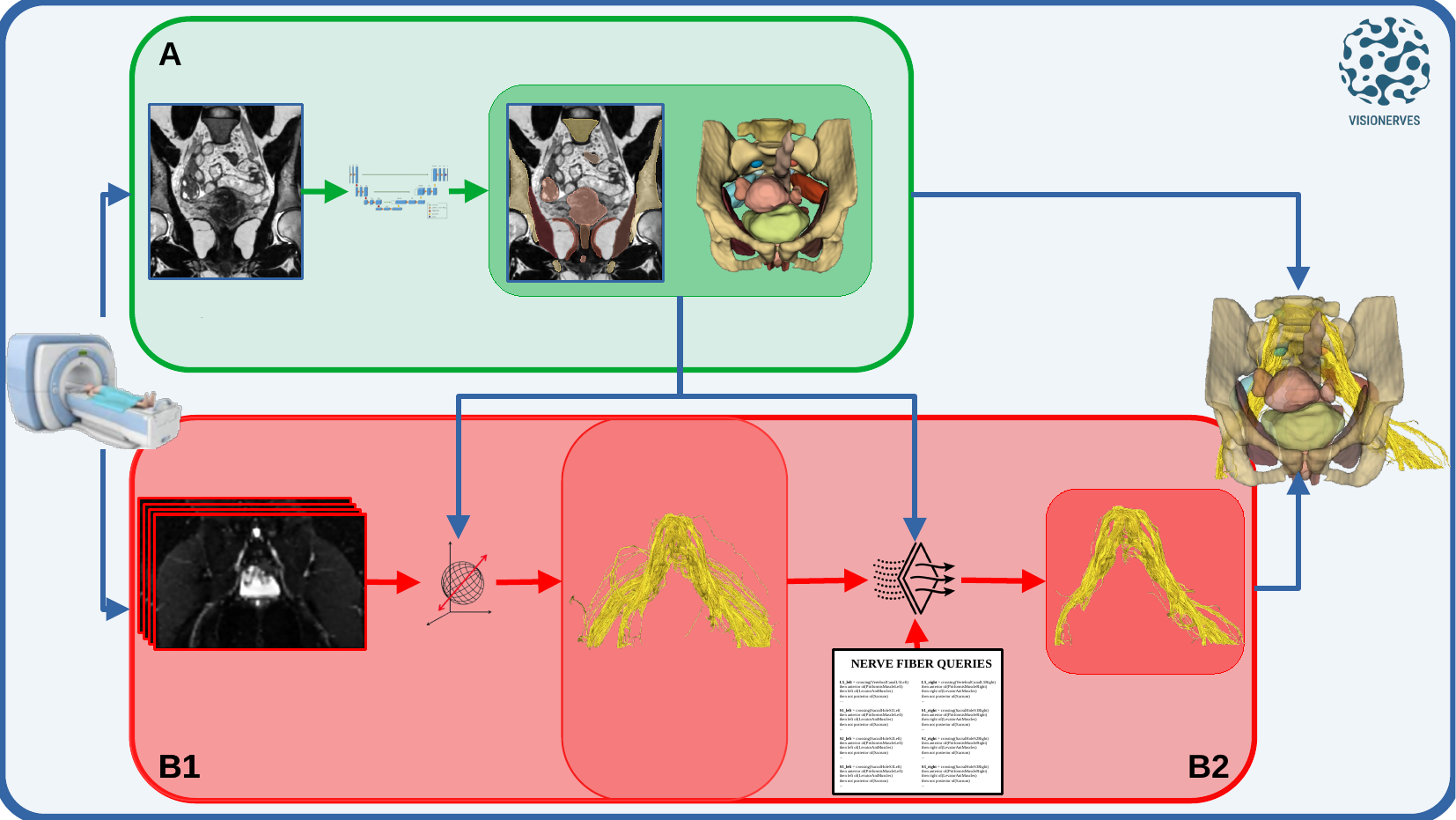}
    \caption{The Visionerves framework (blue) from image acquisition on the left to the final 3D model merging anatomical structures and nerve fibers on the right. The method is composed of: phase A (green)~- Anatomical Structure Segmentation - where a U-Net-based algorithm takes as input a morphological (e.g. T2-w) MR image and outputs an anatomical segmentation; phase B (red)~- Peripheral Nerves Reconstruction (B1) and Recognition (B2) - where a tractography algorithm takes as input a DW image (together with output segmentation of phase A), and outputs a fibers reconstruction, which in turn is passed as input (together with nerve fibers queries and result of phase A) to our symbolic AI filtering method for recognizing the targeted nerves.}
    \label{fig:pipeline}
\end{figure}

\section{The Visionerves Method}
\label{sec:visionerves}
The proposed Visionerves method consists of two primary phases (Figure~\ref{fig:pipeline}): (A) anatomical structure segmentation  from a morphological (e.g. T2-w) MR image (see Section~\ref{sec:A}); 
and (B) peripheral nerves reconstruction and recognition from multi-gradient DW image, leveraging the result of phase A (see Section~\ref{sec:B}).
As described before, a final 3D model can be created, merging the extracted nerve information with the anatomical segmentation. 

\subsection{Phase A - Anatomical Structure Segmentation} 
\label{sec:A}
A U-Net-based algorithm inspired by the well-known nnU-Net~\cite{Isensee2021} and its derivatives (nnDetection~\cite{Baumgartner2021nndetection} for localization and nnInteractive~\cite{Isensee2025nninteractive} for error correction) is employed for rapid and fully automated modeling of anatomical structures from morphological MRI images, such as T2-w or T1-w. 
A preprocessing step (to have a standard dataset representation as in~\cite{Isensee2021}) includes a reorientation of the images in the RAS coordinate system (left to right, posterior to anterior, inferior to superior), a correction of field heterogeneity, and a resampling to a common voxel size. 
The segmentation pipeline includes three custom U-Net subsystems: one for localization (bounding box determination), one for semantic segmentation, and one for error correction via user interaction.  

This phase A of the pipeline enables fast and user-friendly 3D anatomical model generation, avoiding manual segmentation, however it is not central to our method and it can be easily replaced with alternative state-of-the-art approaches. For these reasons we will not elaborate on it further. 
Nevertheless it is important to mention that in addition to enabling the functioning of phase B, this part also allows for the visualization of the nerves in their anatomical context, leading to a better understanding of their relationships to the different anatomical structures. 

\subsection{Phase B - Peripheral Nerves Reconstruction and Recognition}
\label{sec:B}
For the identification of the PNS, in contrast to previous works~\cite{Smith2012,Wassermann},  we propose to represent and formalize anatomical knowledge, usually given in natural language, in first order logic, with the associated syntactic reasoning abilities. Taking inspiration from~\cite{Delmonte}, we also propose to associate it with fuzzy semantics. Spatial relations between nerves and anatomical structures play an important role in the description of nerves. They are represented as predicates in the logic, for which a degree of satisfaction is computed using mathematical morphology and fuzzy sets~\cite{BlochRalescu}. These relations include distances, directions and connectivity with respect to segmented structures from phase A. This hybrid approach combines formal descriptions (syntactic part) with concrete representations in the spatial domain and as degrees of satisfaction taking values in $[0,1]$ (semantic part).
Fuzzy representations hence inherently solve the semantic gap, establishing links between abstract clinical concepts and image information. 

\subsubsection{Phase B1 - Automatic Tractography for Reconstruction.}
\label{sec:B1}
The first part of phase B consists of fiber reconstruction via a tractography algorithm~\cite{Tournier} applied to the multi-gradient DWI image. 
A preprocessing (as in~\cite{Tournier}) includes denoising, Gibbs artifact removal, eddy currents and motion corrections, and correction of field heterogeneity. Following this, a tractography algorithm is applied (algorithm choice left to the user depending on DWI parameters and the studied body region). In order to narrow the reconstruction space 
as well as make this process automatic and reproducible, ROIs for seeding and inclusion zones (which are obligatory to be crossed in the specified order) are produced using either directly the segmented anatomical structures or by using spatial relations to define regions where they are satisfied (e.g. region ``anterior of structure A'' AND ``to the right of structure B''). In this part, spatial relations are binarized, in order to be used in tractography algorithms. In addition, the created regions are large enough to not limit the reconstruction space too much, given the potential deviations from normal anatomy.
It is important to note that, in order to have a good voxel matching between anatomical structures (produced from the MRI) and DWI, a registration might be necessary.

\subsubsection{Phase B2 - Symbolic AI for Recognition.}
\label{sec:B2}
Once the tractogram is reconstructed, we perform a filtering to recognize only the nerve fibers, differentiating them from muscle and tissue fibers as well as potential noise. 
The medical knowledge encoded in the logic is translated into queries used by the recognition algorithm.
For each nerve bundle, a query representing its anatomical path is created using the spatial relations, with the possibility of combining the relations with AND/OR operators, creating exclusion zones with NOT operators, and ordering the nerve segments with THEN (i.e. ``sequential" AND) operators. 
Once the query is built, its degree of satisfaction is assessed at each point along the fiber by mapping every point to the corresponding defined fuzzy regions.
A fiber is considered to satisfy a query if it sequentially validates all specified spatial relations, where to validate means that the average of the non-zero fuzzy values along the fiber is higher than a specified threshold. All fibers that fulfill these criteria are then aggregated to form a bundle representing the targeted nerve. More details on the fuzzy logic modeling and on phase B2 can be found in~\cite{FuzzIEEE}.

\section{Application on the pelvic region for endometriosis}
\label{sec:app}
We applied our Visionerves method on the lumbosacral plexus in the 10 endometriosis cases, for whom a nerve fiber analysis could be a strong aid in the understanding of this disease as explained in Section~\ref{sec:intro}. Furthermore, since this is a highly innervated region with different muscles and organs, it represents an ideal subject of application for our method.

\subsection{Database}
\label{sec:data}
For the segmentation system of phase A, we used 168 T2-w MRI images of 131 patients (ranging from 2 months old to 20 years old) belonging to a proprietary database licensed by the Hôpital Necker-Enfants malades of Paris), with reference images manually annotated by expert surgeons and radiologists over the course of several years using 3DSlicer software~\cite{Fedorov2012Slicer}. Pelvic bones (L5 vertebra, hip bones and sacrum), muscles (piriformis, obturator and levator ani), visceral organs (bladder, colon and rectum) and reproductive organs (ovaries, uterus and vagina) were labeled, in addition to other regions of interest (sacral foramina from S1 to S3, sacral canal, intervertebral foramina of L5) facilitating nerve detection.

We then applied the complete Visionerves method (phases A and B) on 10 different adult female patients (5 diagnosed and 5 suspected endometriosis, both groups in an age range from 20 to 50 years old), gathered at Hôpital européen Georges-Pompidou of Paris. For each patient, a couple of T2-w MRI (reconstruction voxel size 0.5$\times$0.47$\times$0.47 $mm^3$) and multi-gradient DWI image (acquisition voxel size 3.3$\times$2.3$\times$3.6 $mm^3$, NEX 1, 50 directions, b-value 600) was acquired in a 3T GE Signa Architect machine during a preliminary research (for a future prospective study) and used retrospectively after anonymization. Nerve reference reconstructions were created using the method of phase B1 plus the use of a ROI mask, in order to select the fibers passing through, exclusively and completely, within it. This further constrains the search area to the region we consider to represent the true pathway of fiber passage. Such a ROI mask was produced under the supervision of expert surgeons and radiologists via manual segmentation of tubes enclosing each bundle of nerve fibers (given the difficulty of accurately segmenting these structures). 
All the manual annotations detailed in this paragraph were performed using 3DSlicer software~\cite{Fedorov2012Slicer}.

Finally the nerve queries were written with the help of clinical experts, leveraging anatomy books~\cite{Gray}, literature~\cite{Manganaro2014} and knowledge, and aiming to make them generalizable across different cases.
For example a query for recognizing the left S2 nerve is (the parameters defining the relations and threshold values are not mentioned here for the sake of readability):

\smallskip

\noindent
\begin{tabular}{lp{12cm}}
S2\_left $=$ & crossing(SacralHoleS2Left) then  anterior\_of(PiriformisMuscleLeft)\\
& then left\_of(LevatorAniMuscles) then not posterior\_of(Sacrum)\\
& then not (crossing(SacralHoleS1Left) or crossing(SacralHoleS3Left)) \\
& then not left\_of(PiriformisMuscleLeft) \\
& then not anterior\_of(ObturatorMuscleLeft)\\
& then not between(ObturatorMuscleLeft, ObturatorMuscleRight)
\end{tabular}

\subsection{Results and Discussion}
The networks in phase A were implemented from scratch using Tensorflow 2.16. We used 89 T2-w images for training, 16 as validation set and 63 as test set. All images were preprocessed as described in Section~\ref{sec:A} with a common voxel size of 0.88$\times$0.88$\times$0.88 $mm^3$. The method showed high quality segmentation of the pelvic structures described in Section~\ref{sec:data} with Dice indices exceeding 85\% for dense structures and Average Surface Distance less than 2~mm for elongated or small structures. The pelvic region was consistently localized by the first network and the U-Net-based error correction proved effective for minor pelvic structures, providing satisfying results for clinicians in around 2 minutes in worst case scenarios. Although the error correction phase reduces the level of automation of phase A, it serves to decouple potential segmentation errors from directly propagating into the results of phase B.

\begin{table}[h!]
\caption{Quantitative results of phase A of the Visionerves method in average (and standard deviation) for 10 endometriosis cases on different anatomical structures using Dice and Average Symmetric Surface Distance (ASSD). Results are shown before the use of the U-Net-based error correction. Bones are L5 vertebra, hip bones and sacrum; muscles are
piriformis, obturator and levator ani; visceral organs are bladder, colon and rectum; reproductive organs are ovaries, uterus and vagina; specific ROIs are sacral foramina from S1 to S3, sacral canal and intervertebral foramina of L5.}
\centering
\begin{tabulary}{\textwidth}{|L|C|C|C|C|C|}
\hline
Structure    & Bones & Muscles & Visceral organs & Reproductive organs & Specific ROIs \\
\hline
Dice {[}\%{]} & 95.7 & 91.3 & 89.1 & 86.4 & 86.5 \\
& (0.14) & (0.54) & (0.92) & (0.78) & (1.26) \\
\hline
ASSD {[}$mm${]}  & 0.22 & 0.36 & 0.97 & 0.93 & 0.43 \\       
 & (0.09) & (0.42) & (1.18) & (0.62) & (0.51)  \\  
\hline
\end{tabulary}
\label{tab:ResultsA}
\end{table}

\begin{figure}[h!]
    \centering    \includegraphics[width=0.9\textwidth]{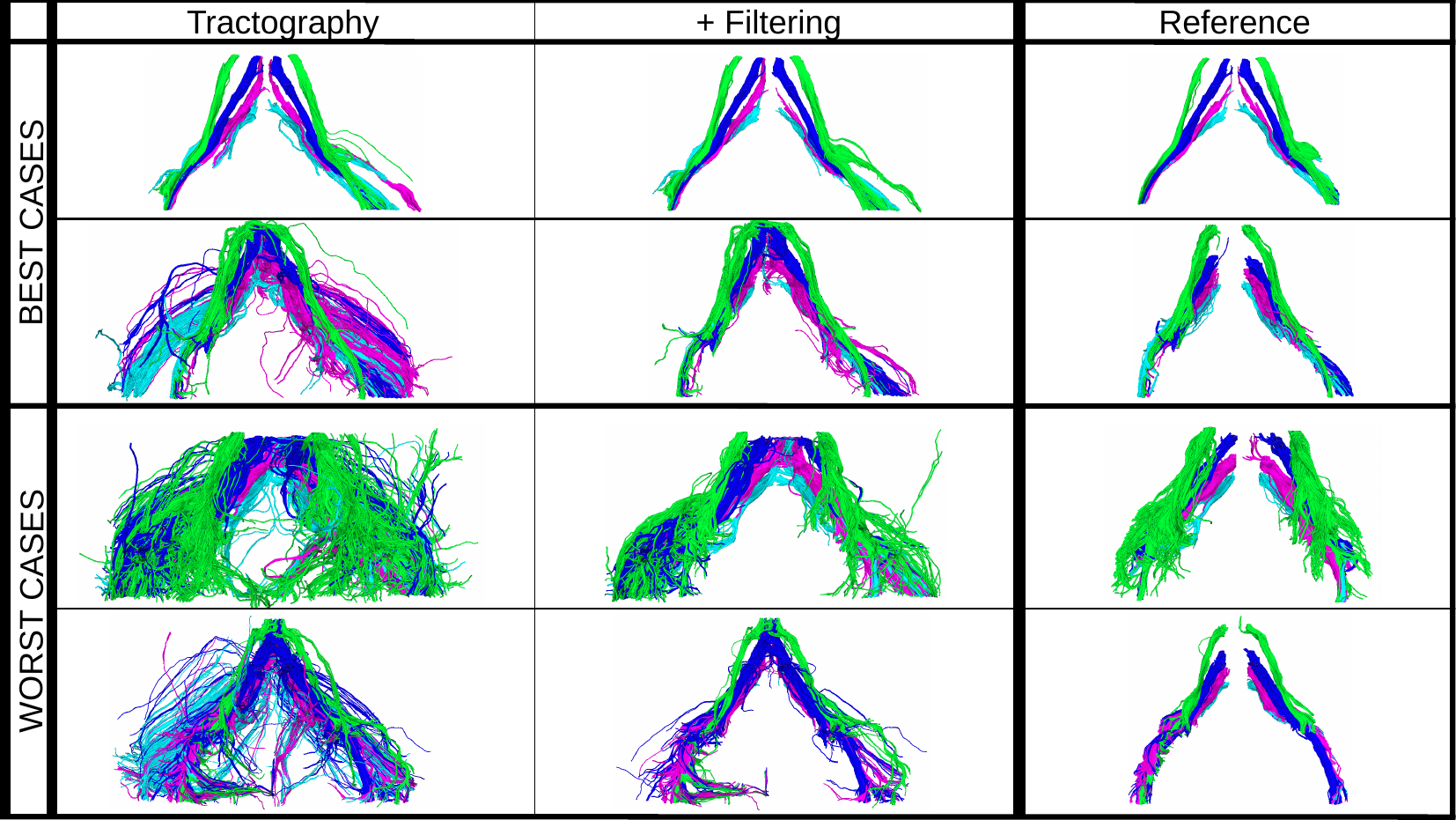}
    \caption{Qualitative results of the Visionerves method (before and after recognition filtering) for the two best cases and two worst cases of the 10 endometriosis cases. Nerve reference reconstructions are produced as described in Section~\ref{sec:data}. Color code for nerve bundle: L5 in green, S1 in blue, S2 in magenta and S3 in cyan.}
    \label{fig:results}
\end{figure}

The results of the complete Visionerves method on the 10 subjects with both T2-w and DWI acquisitions are shown in Table~\ref{tab:ResultsA} for phase A (segmentation). These results are reported for completeness, even though segmentation is not the central focus of our method; they are shown before error correction, which was rarely necessary. Phase B results (reconstruction and recognition) are shown in Table~\ref{tab:Results} and were obtained using the 10 nerve reference reconstructions available from our patient cohort. After preprocessing, nerve bundles were extracted using raw tractography. The sacral or intervertebral foramen corresponding to each nerve (for each side) served as the seed labelmap, and fiber selection was constrained to those containing at least one point within the region traversed by the sciatic nerve (where all the four analyzed fiber bundles are confluent, see Figure~\ref{fig:results} for a better understanding). This region was constructed using the binarized spatial relations that we defined in Section~\ref{sec:B1}. This phase was executed using MRTrix3 software~\cite{Tournier} and we used a FOD-based algorithm with deterministic tracking, called ``SD STREAM"~\cite{Tournier2011}, with minimum FOD amplitude for seeds of 0.15, FOD cut-off of 0.10, maximum angle of 45 degrees and step size of 3~$mm$. This raw tractogram is referred as just ``Tractography'' in Table \ref{tab:Results}, and represents the current state of the art in PNS nerve recognition for traditional methods. Learning-based approaches could not be evaluated, primarily due to the challenges associated with detailed manual nerve segmentation in the T2-w images, and, more critically, the infeasibility of training a neural network given the only 10 cases with DWI and the lack of publicly available pre-trained models.
We then applied our filtering method based on symbolic AI using the nerve queries defined as in Section~\ref{sec:data} (see also Figure~\ref{fig:pipeline}) to the raw tractograms produced. These results are referred as ``+ Filtering'' in Table \ref{tab:Results}. 
Since the Dice score is not well adapted to the thin and tubular structure of the nerves, we also used the precision score (recall was not considered, as it cannot exceed the performance achieved by tractography) and multiple distance metrics that are the Average Symmetric Surface Distance (ASSD), the Average Symmetric Centerline Distance (ASCD) and the Absolute Length Difference (ALD). In order to make these measurements, each fiber bundle was transformed into a single labelmap.

\begin{table}[h!]
\caption{Quantitative results of the Visionerves method (pre- and post-filtering) in average (and standard deviation) for 10 endometriosis cases on 4 different lumbosacral nerves (divided by sides) using Dice, precision, Average Symmetric Surface Distance (ASSD), Average Symmetric Centerline Distance (ASCD) and Absolute (Euclidean) Length Difference (ALD).}
\centering
\begin{tabulary}{\textwidth}{|C|N|C|C|C|C|C|}
    \hline
    Nerve Bundle & Visionerves & Dice [\%] & Precision [\%] & ASSD [mm] & ASCD [mm] & ALD [mm] \\
    \hline
    \multirow{2}{2em}{\\L5 left} & Tractography & 49.56 (19.22) & 35.09 (18.69) & 9.66 (4.42) & 31.67 (19.33) & 30.64 (33.73) \\ \cline{2-7}
    & + Filtering & \textbf{70.70 (16.26)} & \textbf{63.65 (17.35)} & \textbf{5.02 (4.38)} & \textbf{14.27 (16.11)} & \textbf{26.49 (30.52)} \\
    \hline
    \multirow{2}{2em}{\\L5 right} & Tractography & 55.81 (14.29) & 40.21 (14.21) & 6.30 (3.63) & 18.28 (12.53) & \textbf{22.25 (27.78)} \\ \cline{2-7}
    & + Filtering & \textbf{64.86 (9.85)} & \textbf{58.74 (14.31)} & \textbf{3.49 (2.38)} & \textbf{10.01 (8.89)} & 24.98 (30.65) \\
    \hline
    \multirow{2}{2em}{\\S1 left} & Tractography & 56.37 (21.52) & 42.37 (21.95) & 6.02 (3.48) & 14.14 (8.34) & 36.12 (22.75) \\ \cline{2-7}
    & + Filtering & \textbf{74.31 (11.95)} & \textbf{70.02 (16.47)} & \textbf{2.37 (1.80)} & \textbf{7.20 (7.00)} & \textbf{21.37 (23.47)} \\
    \hline
    \multirow{2}{2em}{\\S1 right} & Tractography & 60.40 (24.97) & 47.79 (27.36) & 6.59 (4.39) & 13.44 (11.44) & \textbf{14.98 (18.03}) \\ \cline{2-7}
    & + Filtering & \textbf{74.46 (19.94)} & \textbf{69.65 (25.02)} & \textbf{2.12 (2.54)} & \textbf{5.81 (5.32)} & 15.03 (18.34) \\
    \hline
    \multirow{2}{2em}{\\S2 left} & Tractography & 50.96 (18.43) & 36.57 (20.23) & 7.27 (3.11) & 19.30 (15.20) & 35.76 (35.40) \\ \cline{2-7}
    & + Filtering & \textbf{69.32 (12.48)} & \textbf{59.45 (17.39)} & \textbf{3.52 (2.27)} & \textbf{11.95 (18.01)} & \textbf{32.40 (27.03)} \\
    \hline
    \multirow{2}{2em}{\\S2 right} & Tractography & 56.78 (17.69) & 41.83 (17.67) & 6.37 (4.11) & 13.54 (8.82) & 25.30 (17.36) \\ \cline{2-7}
    & + Filtering & \textbf{70.71 (10.05)} & \textbf{61.19 (12.66)} & \textbf{3.01 (2.61)} & \textbf{6.37 (5.75)} & \textbf{17.77 (17.68)} \\
    \hline
    \multirow{2}{2em}{\\S3 left} & Tractography & 44.54 (18.73) & 30.55 (16.40) & 7.48 (3.32) & 17.47 (7.07) & 37.68 (32.23) \\ \cline{2-7}
    & + Filtering & \textbf{54.78 (15.95)} & \textbf{56.90 (21.35)} & \textbf{3.19 (1.83)} & \textbf{7.14 (2.82)} & \textbf{30.61 (21.72)} \\
    \hline
    \multirow{2}{2em}{\\S3 right} & Tractography & 39.49 (18.56) & 26.32 (15.11) & 9.66 (4.40) & 21.69 (12.07) & 33.70 (27.81) \\ \cline{2-7}
    & + Filtering & \textbf{54.06 (18.77)} & \textbf{55.26 (23.43)} & \textbf{4.28 (3.54)} & \textbf{6.88 (2.67)} & \textbf{28.55 (22.80)} \\
    \hline
\end{tabulary}

\label{tab:Results}
\end{table}

Results show that just using tractography, even guided via ROIs for seeding and inclusion, is not enough to eliminate all spurious fibers. 
By contrast, the proposed approach via symbolic AI allows Dice results to be increased by at least 15$\%$ and precision by at least 25$\%$ for most fibers, indicating proper elimination of false positives with nearly no loss of true positives. Distance metrics (more suitable for evaluation of elongated structures) decrease drastically to half a centimeter for the entire bundle surface and a centimeter when considering the centerline. 
The difference in length remains around 2~cm, indicating that some fibers are not tracked all the way to the end of the portion of the fiber associated with the sciatic nerve.
According to clinical experts, these distances are acceptable (to a certain extent) for clinic use, in particular when examined along with a visual assessment. Notably, Dice and precision scores are lower for S3 due to its smaller fiber diameter, making these evaluation measures less suitable; however, distance-based measures remain comparable to those of other nerve bundles.

Some qualitative results are presented in Figure~\ref{fig:results}: for the two best cases, except for a few spurious fibers, the bundles overlap almost perfectly, while in the two worst cases most false positives are still effectively filtered (moreover on one case reference nerves are also highly disorganized). Only one case (third row) appears anatomically incorrect, but the reference nerves are also very disorganized. Interestingly, the two best-performing cases correspond to patients with confirmed endometriosis, whereas the worst-performing cases involve patients with suspected endometriosis (who exhibited symptoms of menorrhagia and pelvic pain). It remains unclear whether these suboptimal results are due to the choice of the tractography algorithm, its parameter settings or to intrinsic limitations of the imaging data. A more in-depth study needs to be conducted. Moreover, the filtering parameters and threshold values may probably need to be adjusted in cases where the raw tractogram already appears very disorganized.

\section{Conclusion and Perspectives}
\label{sec:conclusion} 

In this work, we introduced Visionerves, a novel hybrid AI framework that leverages anatomical knowledge for the automatic and reproducible recognition of peripheral nerves. By integrating fuzzy spatial reasoning with symbolic AI and multi-modal MRI, our method addresses key limitations of traditional tractography, most notably its reliance on manual intervention and lack of reproducibility. 
For a preliminary assessment we applied Visionerves in the pelvic region on 10 endometriosis cases. Results demonstrated significant improvements over conventional tractography approaches confirming that our method can substantially reduce spurious fibers while maintaining the structural integrity of the nerve reconstructions. 

By enabling the reproducible and individualized extraction of nerve bundles, Visionerves paves the way for assessing nerve-specific diffusion and morphological characteristics. 
This is particularly valuable in endometriosis, where these fiber properties may enable diagnosis in suspected cases of nerve involvement without the need for surgical biopsy. The S2 and S3 would be the principal nerves studied for this pathology since they are part of the pudendal plexus that is innervating the external genitalia and the perineum. Even though L5 and S1 further constitute the sciatic nerve, they are still of interest due to their proximity to the uterus and ovaries. Other conditions with possible nerve involvement may also benefit from our method (e.g. pelvic cancers), where Visionerves can also be used for pre-surgical planning and longitudinal follow-up.

In order to reach clinically acceptable results, in future work we plan to: (i)~extend the database, including also healthy subjects and other pathologies, in order to enable more informed selection of hyperparameters during both phase B stages; (ii)~refining query formalization (e.g. new spatial relations), and automate its generation from textual anatomical descriptions and its parametrization in cases where the raw tractogram is full of spurious fibers; (iii)~exploring alternative validation criteria for fiber selection or adopt a solution where threshold values are adjusted based on a general “anatomical coherence score” that can be adjusted by the user, such as in~\cite{Delmonte}.
Ultimately, we will test Visionerves on other pelvic nerve fibers (e.g. S4, pudendal, obturator) and other regions, (e.g. skull base, head and neck, brachial plexus), where conventional tractography remains particularly challenging. Once an anatomical segmentation of a region and the trajectories of the nerves in that region are defined, the Visionerves method can be applied in a straightforward manner, leveraging the existing comprehensive set of spatial relations, which can be easily expanded if needed.

\begin{credits}
\subsubsection{\ackname} This work has been funded and supported by Ligue contre le cancer, Fondation Béatrice Denys and Prématuration IP Paris.  This work was performed using HPC Jean-Zay resources from GENCI–IDRIS (Grant 2025-AD011015418). We would like to thank Fatiha Tacine for her assistance with the organization and retrieval of the acquisitions at the Hôpital européen Georges Pompidou. We would like to thank also Alice Sorrentino for her assistance with the organization and segmentation of the acquisitions at the Hôpital Necker-Enfants malades.

\subsubsection{Data Use Declaration.} The 131 patients used in phase A were included under a license granted by the Hôpital Necker-Enfants malades for acquisitions during protocol n°2015-101705-44. The 10 endometriosis patients used for testing the Visionerves framework were included under preliminary acquisitions of a future research clinical protocol n°2024-100538-39 approved by the Hôpital européen Georges-Pompidou.

\subsubsection{\discintname} The authors declare that a patent application has been filed, related to the method presented in this paper. This disclosure is made in the interest of transparency and does not affect the integrity or objectivity of the research. 
\end{credits}

\bibliographystyle{splncs04}
\bibliography{refs.bib}

\end{document}